\documentclass[journal]{IEEEtran}
\usepackage{amsmath,amsfonts}
\usepackage{algorithmic}
\usepackage{algorithm}
\usepackage{array}
\usepackage[caption=false,font=normalsize]{subfig}
\usepackage{textcomp}
\usepackage{stfloats}
\usepackage{url}
\usepackage{verbatim}
\usepackage{amssymb}
\usepackage{bbm}
\usepackage{multirow}
\usepackage{booktabs}
\usepackage{graphicx}
\usepackage{cite}
\usepackage{xcolor}
\definecolor{linkblue}{rgb}{0.2, 0.4, 0.8} 
\usepackage[colorlinks=true, linkcolor=linkblue, citecolor=linkblue, urlcolor=magenta]{hyperref}

\usepackage{cleveref}  % For \cref command, MUST BE AFTER hyperref
\makeatletter
\long\def\@makecaption#1#2{%
  \vskip\abovecaptionskip
  \sbox\@tempboxa{#1: #2}%
  \ifdim \wd\@tempboxa >\hsize
    #1: #2\par
  \else
    \global \@minipagefalse
    \hb@xt@\hsize{\hfil\box\@tempboxa\hfil}%
  \fi
  \vskip\belowcaptionskip}
\makeatother

\crefname{figure}{Fig.}{Figs.}
\Crefname{figure}{Fig.}{Figs.}
\crefname{table}{TABLE}{TABLES}
\Crefname{table}{TABLE}{TABLES}
\crefname{section}{Section}{Sections}
\Crefname{section}{Section}{Sections}
\crefname{equation}{}{} % No "Eq." prefix for equations, just the number in parens
\newcommand{\rev}[1]{\textcolor{black}{#1}}
\usepackage{eso-pic} % <-- Add this line

\begin{document}

\title{In-Field Mapping of Grape Yield and Quality with Illumination-Invariant Deep Learning}

\author{Ciem Cornelissen, Sander De Coninck, Axel Willekens, Sam Leroux, Pieter Simoens~\IEEEmembership{}
        % <-this % stops a space
\thanks{Manuscript received July 31, 2025. \textit{(Corresponding author: Ciem Cornelissen.)}}
\thanks{C. Cornelissen, S. De Coninck, S. Leroux, and P. Simoens are with IDLab, Department of Information Technology at Ghent University - imec (e-mail: ciem.cornelissen@ugent.be; sander.deconinck@ugent.be; sam.leroux@ugent.be; pieter.simoens@ugent.be).}
\thanks{A. Willekens is with Ghent University, IDLab ‐ Imec ‐ AI and Robotics Lab (AIRO), Zwijnaarde, Belgium (e-mail:Axel.Willekens@ilvo.vlaanderen.be).}
\thanks{A. Willekens is with Institute for Agriculture, Food and Fishery Flanders (ILVO),
Merelbeke‐Melle, Belgium (e-mail:Axel.Willekens@ilvo.vlaanderen.be).}}


% \IEEEpubid{\begin{minipage}{\columnwidth}\ \\[24pt] 
% % \centering
%   © 2025 IEEE. Personal use of this material is permitted. Permission from IEEE must be obtained for all other uses, in any current or future media, including reprinting/republishing this material for advertising or promotional purposes, creating new collective works, for resale or redistribution to servers or lists, or reuse of any copyrighted component of this work in other works.
% \end{minipage}}

\AddToShipoutPictureBG*{%
  \AtPageLowerLeft{%
    \put(50pt,83pt){\parbox[t]{\columnwidth}{%
      \footnotesize © 2025 IEEE.  Personal use of this material is permitted.  Permission from IEEE must be obtained for all other uses, in any current or future media, including reprinting/republishing this material for advertising or promotional purposes, creating new collective works, for resale or redistribution to servers or lists, or reuse of any copyrighted component of this work in other works.}}%
  }%
}

\maketitle

% \IEEEpubidadjcol

\enlargethispage{-4\baselineskip} 

\begin{abstract}
This paper presents an end-to-end, IoT-enabled robotic system for the non-destructive, real-time, and spatially-resolved mapping of grape yield and quality (Brix, Acidity) in vineyards. The system features a comprehensive analytical pipeline that integrates two key modules: a high-performance model for grape bunch detection and weight estimation, and a novel deep learning framework for quality assessment from hyperspectral (HSI) data. A critical barrier to in-field HSI is the ``domain shift" caused by variable illumination. To overcome this, our quality assessment is powered by the Light-Invariant Spectral Autoencoder (LISA), a domain-adversarial framework that learns illumination-invariant features from uncalibrated data. We validated the system's robustness on a purpose-built HSI dataset spanning three distinct illumination domains: controlled artificial lighting (lab), and variable natural sunlight captured in the morning and afternoon. Results show the complete pipeline achieves a recall (0.82) for bunch detection and a $R^2$ (0.76) for weight prediction, while the LISA module improves quality prediction generalization by over 20\% compared to the baselines. By combining these robust modules, the system successfully generates high-resolution, georeferenced data of both grape yield and quality, providing actionable, data-driven insights for precision viticulture.

\end{abstract}


\section{Introduction}
\IEEEPARstart{P}{recision} viticulture (PV) aims to manage the inherent spatial and temporal variability within vineyards to optimize grape quality and yield. Traditional methods for assessing key parameters like sugar content (Brix) and acidity rely on manual, destructive, and labor-intensive sampling \cite{laborIntensive}. These methods are not scalable and fail to provide the high-resolution, real-time data needed for data-driven management. The Internet of Things (IoT), combined with agricultural robotics, presents an opportunity to overcome these limitations. By deploying intelligent, mobile sensing platforms, it becomes possible to automate data collection and generate spatially-resolved data of crop health, enabling targeted interventions and advancing sustainable agriculture.

Central to such an IoT system is the choice of sensor technology. 
Hyperspectral imaging (HSI) is a sophisticated non-destructive technique that allows for the direct quantification of internal quality parameters like Brix and acidity in grapes \cite{precisionAgri}. Crucially, this assessment is performed in-situ by analyzing the grapes' spectral signatures, leaving the fruit intact on the vine. This offers a vital improvement over traditional wet chemistry methods \cite{grape_1, grape_2}, which require berries to be physically removed, crushed, and analyzed, a destructive and labor-intensive process. The high spectral resolution of HSI reveals these chemical signatures \cite{mypaper,chemicalcheese} which are invisible to the human eye or standard RGB cameras \cite{HSIreview}. Integrating HSI onto a robotic platform therefore creates a powerful mobile node for continuous, in-field quality assessment.

However, the practical deployment of HSI-based robotic systems in outdoor environments is hindered by a fundamental challenge: variability in illumination. Hyperspectral sensors are highly susceptible to changes in ambient light caused by the sun's position, atmospheric conditions, and shadows \cite{lightingConditions}. These environmental factors cause significant distribution shifts in the acquired data, a problem known in machine learning as ``domain shift" \cite{labToField}. Consequently, a model trained under one condition (e.g., in a controlled lab or on a sunny morning) fails to generalize when deployed under different lighting (e.g., on a cloudy afternoon), severely limiting the system's reliability and real-world applicability \cite{realWorldChallenges}.

Traditional methods for tackling this issue rely on explicit radiometric calibration, typically by capturing an image of a white reference standard (e.g., a Spectralon® panel) before each scan \cite{whiteReference}. While effective in static settings, this approach is logistically burdensome and infeasible for an autonomous IoT system designed for continuous operation. Requiring a robot to constantly stop, deploy, and measure a clean, perfectly positioned reference panel is impractical and defeats the purpose of high-throughput automation \cite{whiteRefLimitations}. \rev{This highlights a critical gap: the need for an approach that can achieve robustness without relying on these impractical, stop-and-go field calibrations. A key objective of this work is to develop and validate a data-driven framework that learns illumination invariance directly from uncalibrated data, thereby enabling continuous and autonomous in-field operation, which outperforms traditional physics-based correction methods.}

To address the limitations of traditional sampling, this paper presents an end-to-end, intelligent IoT system for acquiring real-time, non-destructive, and spatially-resolved data on grape quality and yield. The primary challenge in developing such a system for field deployment is the data variability caused by changing outdoor illumination. Our work aims to create a system that is not only automated but also robust enough to provide reliable predictions across these dynamic lighting conditions, thereby reducing the reliance on the impractical, stop-and-go field calibrations that hinder continuous robotic operation.

This paper details the development and evaluation of such an intelligent system. The key contributions are:
\begin{itemize}
\item{\textbf{A Novel Illumination-Invariant Model (LISA):} The development and validation of LISA, a domain-adversarial autoencoder that learns to predict grape quality (Brix and Acidity) directly from uncalibrated HSI data. We demonstrate that LISA significantly outperforms baselines by remaining robust to the domain shifts caused by changing natural illumination, removing the need for impractical, frequent field calibration.}
\item{\textbf{A Unique Multi-Domain Validation Dataset:} The creation and use of a purpose-built HSI dataset designed to test domain generalization. This dataset contains paired scans of the same grape bunches under three distinct lighting conditions (controlled lab, morning sun, and afternoon sun), providing a challenging and scientifically rigorous benchmark for model robustness under different lighting conditions \footnote{The dataset is available from the corresponding author upon request.}.}
\item{\textbf{An End-to-End System for Dual Yield and Quality Mapping:} The design, integration, and validation of a complete analytical pipeline that fuses a high-performance yield estimation module (YOLOv11-L + 2D CNN) with our novel LISA framework. This integrated system is the first to demonstrate the generation of comprehensive, georeferenced data of both grape yield and quality from a single, uncalibrated robotic HSI scan.}
\end{itemize}

\section{Related Work}
\subsection{From Lab to Field: Robotic Hyperspectral Sensing for Grape Quality}

The IoT paradigm is central to the modern ``Agriculture 4.0” revolution, enabling a shift towards data-driven management \cite{agriculture4}. In viticulture, this translates to Precision Viticulture, a practice focused on managing in-field variability to optimize resource use and enhance grape quality. A key enabler of PV is the use of autonomous ground vehicles (AGVs) or agricultural robots, which provide the mobility for proximal sensing at the plant or even bunch level \cite{Axel}.

The use of hyperspectral imaging for non-destructive grape quality assessment is well-established, though early research is largely confined to controlled laboratory settings. These seminal studies demonstrated the fundamental capability of HSI to quantify a wide range of internal parameters by correlating spectral data with wet chemistry measurements. For instance, Gabrielli et al. \cite{Gabrielli2023} successfully developed models to predict sugar, acid, and phenolic content across seven grape varieties with high accuracy ($R^2 > 0.81$). Similarly, Lyu et al. \cite{Lyu2024} and Gomes et al. \cite{Gomes2021} focused on predicting Total Soluble Solids (TSS/Brix) and acidity, with the latter showcasing the potential of deep learning for robust predictions across different vintages. Research has also extended to more subtle characteristics, such as the aromatic profiles of grapes, where Marín-San Román et al. \cite{SanRoman2024} used HSI to estimate volatile compounds. While proving the principle, these static, lab-based approaches are not scalable for vineyard-wide management.

The recent trend has thus been to move this technology from the lab into the field by integrating HSI sensors onto mobile robotic platforms. This creates powerful mobile IoT nodes capable of generating the detailed, spatially-resolved data required for PV. However, this transition introduces a significant challenge: contending with the dynamic and uncontrolled outdoor environment.

Recent state-of-the-art studies exemplify this shift and its associated hurdles. For instance, Benelli et al. \cite{hsigrape0} utilized a moving wagon equipped with a push-broom HSI system to predict soluble solids content (SSC) in Sangiovese grapes. Their work focused on the critical challenge of automatically selecting Regions of Interest (ROIs), using a PLS-DA model to discriminate grape pixels, even separating them into `sunny' and `shady' classes, from the complex background. Similarly, Bertoglio et al. \cite{hsigrape2} developed a framework for `on-the-go' ripeness estimation using a snapshot HSI camera on a robotic platform. Their work addressed the technical challenge of correcting for spectral band misalignment caused by vehicle motion and employed a Mask R-CNN model to segment individual grape bunches.

While these studies demonstrate the significant potential of robotic HSI systems, they also underscore a fundamental and persistent obstacle: the impact of variable illumination. Both Benelli et al.'s explicit separation of `sunny' and `shady' pixels and Bertoglio et al.'s reliance on calibration under natural light highlight that inconsistent lighting is a primary source of error. Their methods address this by segmenting the problem and employing standard calibration.

\subsection{Illumination Correction in Hyperspectral Imaging}
Traditional approaches for tackling illumination-induced domain shift focus on recovering an object's intrinsic reflectance through explicit correction methods. The most common is radiometric calibration using a reference panel (e.g., a Spectralon® panel), where the target's radiance is normalized by the reference's \cite{iN_Field_white_reference}. While effective in static environments, this is often impractical in dynamic field settings, as the frequent re-measurement of a reference panel is difficult to automate and logistically burdensome. Other techniques include empirical corrections like Internal Average Reflectance (IAR) and Flat Field Correction, which often lack accuracy \cite{limitationsIAR}. More complex approaches involve physical modeling, such as using radiative transfer codes (e.g., MODTRAN \cite{MODTRAN}) to account for atmospheric effects, or subspace models like LOGSEP \cite{LOGSEP} which approximate illumination and reflectance as linear combinations of basis spectra. However, these methods can be computationally intensive, rely on hard-to-obtain parameters, or cannot recover absolute reflectance, limiting their general applicability \cite{MODTRANlimitations}.

\subsection{Domain Adaptation for HSI Analysis}
The limitations of explicit modeling have motivated a shift towards data-driven deep learning solutions. These methods aim to learn robust representations directly from the data, rather than modeling the physics of light. One common strategy is data augmentation. For instance, Generative Adversarial Networks (GANs) \cite{GAN} have been used to synthesize HSI data, thereby exposing the model to a wider range of variations during training and improving its robustness \cite{dataAug1, dataAug2}.

A more direct approach is to frame the problem as one of domain adaptation or domain generalization. The goal is to learn features that are invariant to the domain (e.g., `lab' vs. `field' illumination) while remaining discriminative for the main task. This paradigm is highly relevant, drawing on foundational work in the broader computer vision community. A key technique is adversarial learning, popularized by Domain-Adversarial Neural Networks (DANN) \cite{DANN}. In this approach, a feature extractor is trained simultaneously to support a primary task and to ``fool" a domain discriminator that tries to identify the data's origin. This forces the features to become domain-agnostic. This principle has been successfully applied to HSI, with frameworks like SiLL-R-GAN \cite{SILLRGAN} using adversarial training for robust classification. Other related concepts include feature disentanglement. A prominent approach is physics-informed learning, where models like Variational Autoencoders (VAEs) explicitly integrate a physical model of light transport into the decoder. This forces parts of the latent space to correspond directly to physical factors of variation, such as illumination angle \cite{physicalInformedVAE}. In contrast, our work pursues a domain-adversarial approach, which treats illumination as an unknown domain to be made invariant to, rather than a known physical process to be explicitly modeled.

\section{System Architecture and Data Acquisition}
\begin{figure}[!t]
\centering
\includegraphics[width=3in]{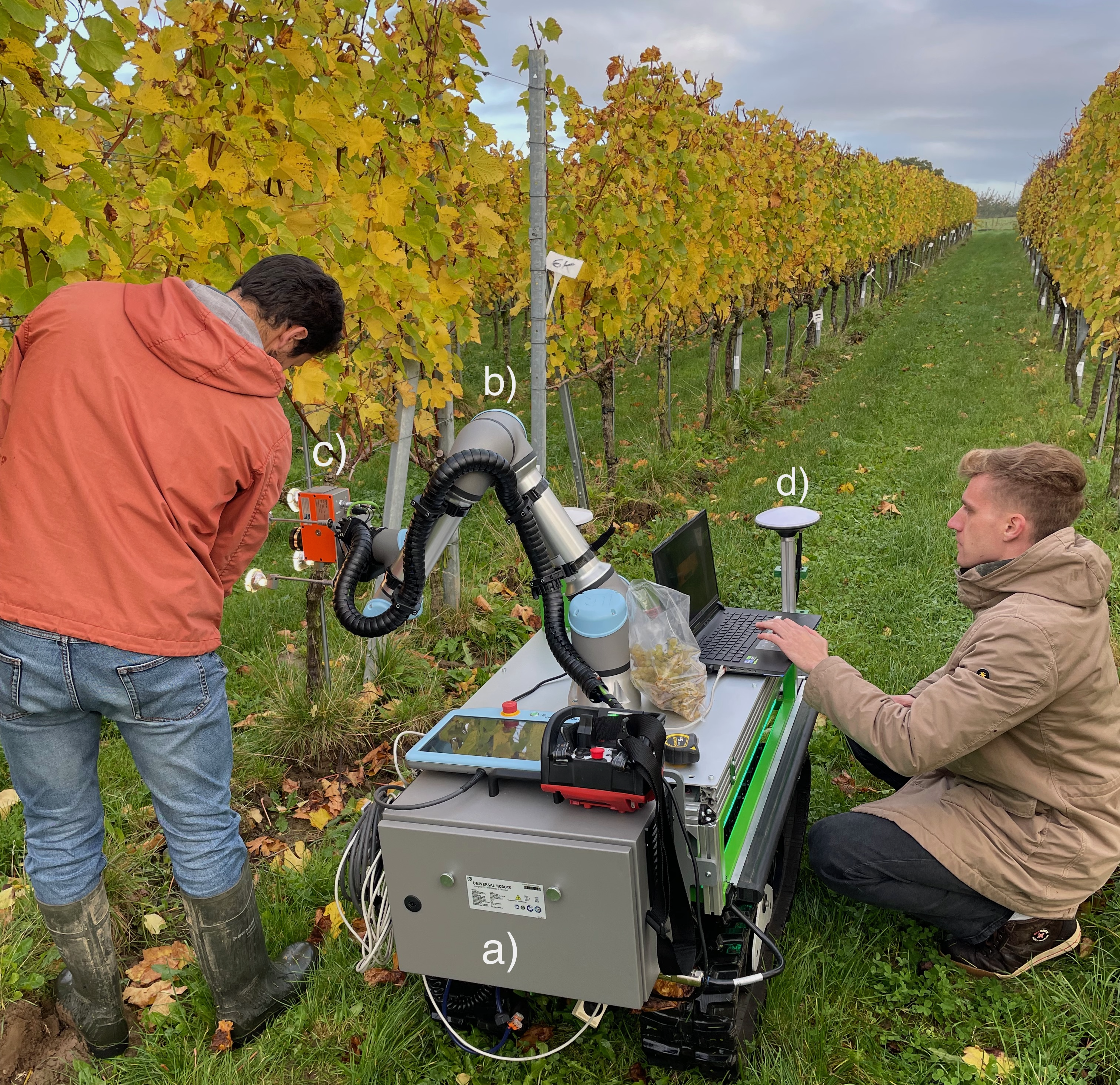}
\caption{The integrated IoT-enabled robotic platform during in-field data acquisition. Key components shown are a) the mobile robotic base, b) a collaborative robotic arm positioning the c) Specim FX10 hyperspectral camera, and d) the GPS antenna for georeferencing.}
\label{fig:robot}
\end{figure}

Our intelligent system is built on a custom-integrated mobile robotic platform designed for autonomous or semi-autonomous navigation and data collection in challenging vineyard environments. \cref{fig:robot} shows the system during an in-field data acquisition campaign. The architecture is modular, consisting of three primary subsystems: the mobile base, the robotic arm for sensor positioning, and the data acquisition and processing unit.

\subsection{System Hardware and Instrumentation}
The robotic platform is built upon the Treebot \cite{Axel}, a robust, all-terrain mobile research platform specifically designed for autonomous tasks in horticultural environments. Its tracked, skid-steer design, originally derived from a mini dumper, provides the necessary stability and maneuverability to traverse uneven ground between vine rows. The electrically-driven base, weighing 267 kg and powered by a 3.2 kWh battery system, serves as a stable foundation for the primary data acquisition hardware. Mounted on this base is a 6-degree-of-freedom (6-DOF) collaborative robotic arm (i.e., a Universal Robots Ur5e). The arm provides the high precision and flexibility required to position the sensor payload at optimal angles and distances from the grapevine canopy, compensating for variations in canopy structure and terrain.

The primary sensor, a Specim FX10 hyperspectral camera \cite{SPECIMFX}, is mounted as the end-effector of the robotic arm. This setup allows the system to perform controlled ``push-broom" scans, where the linear motion of the robot moving along the row provides a spatial dimension of the hyperspectral cube. An integrated high-precision GPS unit, also visible in \cref{fig:robot}, provides geotagging for every captured datapoint, which is crucial for generating the spatially-resolved data. All subsystems are controlled by a distributed onboard computing architecture, mirroring the robust design of the Treebot platform. The high-level processing is handled by an industrial PC, specifically an Intel NUC 10 (NUC10i5FNH). This unit is responsible for running the deep learning modules for yield and quality estimation, as well as managing sensor synchronization and data storage. It interfaces with a Siemens S7-1200 Programmable Logic Controller (PLC), which executes the real-time motion control and low-level commands. This integrated design forms a self-contained IoT node capable of capturing, processing, and georeferencing data directly in the field.

\subsection{Data Acquisition and Dataset Creation}
All hyperspectral data was captured using a Specim FX10 push-broom camera \cite{SPECIMFX}, which senses 1024 spatial pixels across 224 spectral bands in the 400-1000 nm range. The camera was mounted as the end-effector on the robotic arm of our mobile platform. To simulate an operational `on-the-go' scan, full hyperspectral cubes were generated by the linear motion of the robot as it moved along the vineyard rows. The grapes, of the Chardonnay variety, were captured at Proefcentrum Fruitteelt (pcfruit) in Sint-Truiden, Belgium, on October 23rd, 2024.

To specifically test for illumination invariance, our data collection protocol is designed to create a unique multi-domain dataset. The same 60 grape bunches were imaged across all domains, providing paired examples that isolate the effect of illumination from other biological variability. This paired structure supports the validation of our model by isolating the effect of illumination from biological variability during performance evaluation.

\begin{itemize}
    \item \textbf{Lab:} Data captured under stable, consistent artificial lighting ($4\times50$-watt halogen lamps), providing a controlled, high-intensity source domain with a known spectral profile.
    
    \item \textbf{Field-AM (Morning Sun):} Data captured outdoors in the morning. This domain is characterized by a lower solar elevation angle, creating longer shadows and a warmer light spectrum.
    
    \item \textbf{Field-PM (Afternoon Sun):} Data captured outdoors in the afternoon. This second field domain introduces a significant shift from the morning conditions, with a higher solar elevation and a different azimuth. This alters the direction and length of shadows cast by the canopy and within the grape bunches themselves.
\end{itemize}

\begin{figure*}[!t]
    \centering
    \subfloat[The same grape bunch imaged for the three data subsets. Left: Lab subset. Middle: Field-AM subset. Right: Field-PM subset.]{\includegraphics[width=0.47\linewidth]{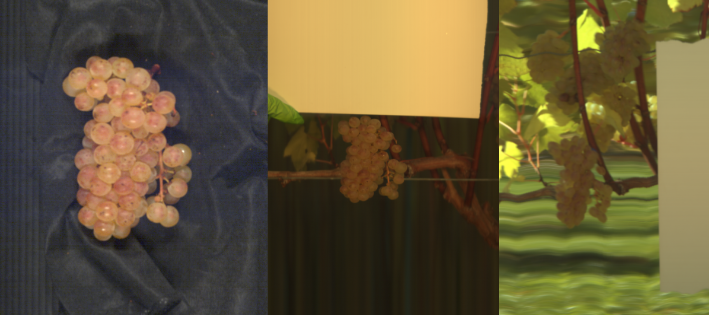}%
    \label{fig:short-a}}
    \hfill
    \subfloat[The average raw spectral intensity (in Digital Numbers, DN) for each subset, with the shaded areas representing one standard deviation. The plot highlights the domain shift between the controlled Lab environment and the two field environments.]{\includegraphics[width=0.49\linewidth]{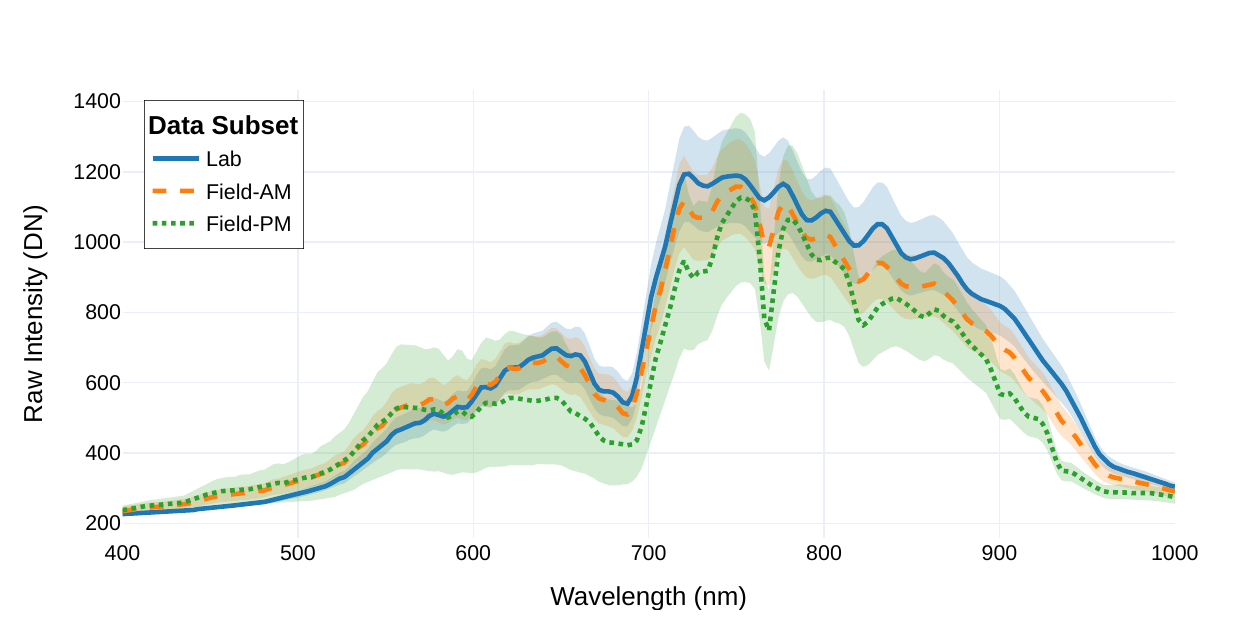}%
    \label{fig:short-b}}
    \caption{Visualization of the pseudo-RGB hyperspectral images and the spectra of different data subsets.}
    \label{fig:short}
\end{figure*}

As illustrated in \cref{fig:short-a}, the visual appearance of the grapes changes dramatically across the domains. This is shown using pseudo-RGB images, i.e., color visualizations synthesized directly from the hyperspectral data cube. Specifically, we create these images by mapping the intensity values from three selected bands to the standard Red, Green, and Blue channels. For our 224-band camera, we map band 114 (centered at $\approx705$ nm) to Red, band 58 ($\approx555$ nm) to Green, and band 20 ($\approx454$ nm) to Blue. The term `pseudo-RGB' is used because this serves as an intuitive, human-readable approximation of color, rather than a true-color photograph from a standard camera.

This shift is quantified in the mean spectra shown in \cref{fig:short-b}. The overall reflectance intensity is highest in the Lab domain and lowest in the Field-PM domain. While all spectra share characteristic features like the vegetation red-edge near 700 nm \cite{redEdge}, the field spectra are distinctly marked by atmospheric effects, such as the oxygen (\ensuremath{\mathrm{O_2}}) absorption band around 760 nm \cite{sunspectrum}. These variations in both intensity and spectral shape constitute the core challenge that robust HSI systems must overcome for practical field deployment.

To train and evaluate the different modules of our analytical pipeline, the dataset is annotated at two distinct levels, corresponding to our primary tasks of quality prediction and yield estimation.
\begin{itemize}
    \item For Quality Prediction (LISA): Our analysis focuses on learning illumination-invariant features from individual grape berries. Therefore, for this task, we extracted spectral signatures by manually annotating small, homogeneous regions on single grapes within the hyperspectral cubes. This process resulted in 360 annotated grape samples for the Lab domain, and 360 samples for each of the Field-AM and Field-PM domains. Each of these annotated samples has a corresponding ground truth measurement for Brix and Acidity, obtained destructively using a digital refractometer and titration, respectively.
    \item For Yield Estimation (YOLO): To train the bunch detection and weight estimation module, we annotated the dataset at the bunch level. In total, 162 images were annotated by drawing bounding boxes around each visible grape bunch. For 76 unique bunches, the ground truth weight is also measured in the lab using a precision scale. This set of 76 bunches is composed of the 60 bunches used in the multi-domain quality study, supplemented with an additional 16 bunches. This data is used to train the YOLO-based object detector \cite{yolo11} and the subsequent weight regression model.
\end{itemize}

\section{Analytical Pipeline and System Integration}
This section details the analytical pipeline that processes the hyperspectral data captured by our robotic platform. The pipeline is designed to operate in real-time on the robot's edge-computing device, integrating two primary deep learning systems: a yield estimation module, which combines a YOLO-based detector with a weight regression model, and our novel LISA framework for robust quality prediction

\begin{figure*}[tbp]
    \centering
    \includegraphics[width=2\columnwidth]{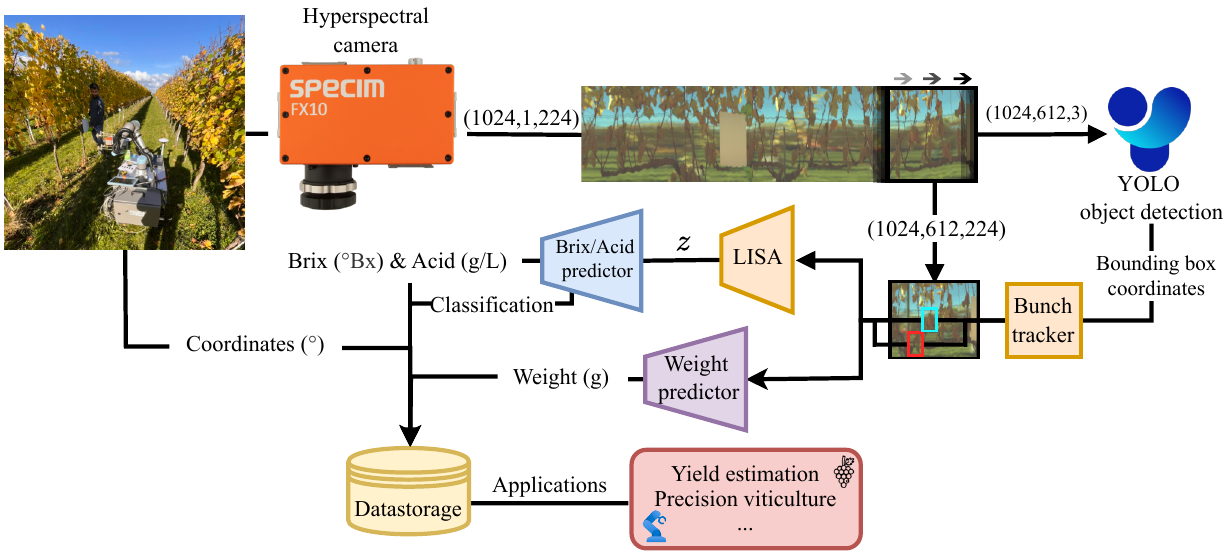}
    \caption{\rev{Overview of the integrated data processing workflow. The robotic platform captures a continuous, georeferenced hyperspectral scan. The system processes this data stream using a sliding window approach. For each window, a pseudo-RGB image is synthesized and passed to a YOLO-based object detector. A bunch tracker identifies unique bunches across consecutive overlapping windows. Once a bunch is fully captured within the frame, its full-spectrum HSI data is extracted and analyzed by two parallel deep learning modules for yield (weight) and quality (Brix/Acidity) prediction. The quality pipeline includes a dedicated preprocessing step before the data is fed into our LISA framework. All per-bunch predictions are georeferenced and aggregated to create spatial data for precision viticulture.
    }}
    \label{fig:workflow_overview}
\end{figure*}

The overall workflow is illustrated in \cref{fig:workflow_overview}. As the robot captures a continuous HSI scan of the vineyard row, the system processes it sequentially. To handle the continuous data stream and ensure objects are analyzed correctly the system uses an overlapping sliding window to ensure bunches are only processed when they are fully in frame. To avoid duplicate predictions, the system uses a bunch tracker to identify unique grape bunches as they move through the camera's field of view. For each unique bunch that is detected and tracked, the following analytical steps are performed:
\begin{enumerate}
    \item \textbf{Bunch Detection and Localization:} A pseudo-RGB image is generated of a sliding window from the HSI data. The YOLO-based model processes this image to identify all visible grape bunches and outputs their bounding box coordinates. The tracker then associates these detections with existing or new bunch identities.

    \item \textbf{Yield Prediction:} Once a tracked bunch is determined to be optimally framed (i.e., centered and fully visible), the hyperspectral data within its bounding box is passed to a dedicated regression model to estimate its weight in grams.

    \item \textbf{Quality Prediction with LISA:} Concurrently, spatio-spectral patches from within the same bounding box are analyzed by the LISA framework. A multi-head architecture, including a classification head, acts as a mask to ensure only grape pixels contribute to the final prediction. The regression heads then infer the average Brix and Acidity for a patch. The prediction for the bunch is the average prediction over all patches classified as grape. 

    \item \textbf{Data Aggregation and Mapping:} The final, georeferenced data point i.e., a tuple containing `\{latitude, longitude, estimated weight, average Brix, average acid\}', is stored for each unique bunch. 
\end{enumerate}

This continuous stream of aggregated data can be used to generate high-resolution spatial variability maps. These maps visualize the distribution of yield, sugar content, and acidity across the vineyard, providing the insights needed for data-driven differential harvesting and other precision management practices.

\subsection{Hyperspectral Data Preprocessing and Input Formulation}

A key objective of our work is to develop a model that is robust to real-world conditions, including in the absence of frequent calibration. Therefore, for the primary input to our LISA framework, we deliberately bypass traditional radiometric calibration and work directly with the raw hyperspectral data. 

The preprocessing, before the LISA component, consists of 
\rev{ A Savitzky-Golay (SG) first derivative filter that is applied directly to the raw spectral data. The optimal parameters for this filter, determined through a hyperparameter search detailed in \cref{hyperparameter}, are a window length of 15 and a polynomial order of 2. This step is crucial for removing baseline shifts and accentuating the subtle absorption features related to chemical composition, which is a standard technique in chemometrics to enhance signal quality.}
The transformed spectral data is then formatted into $8\times 8$ spatio-spectral patches. This results in an input tensor of shape $8\times 8\times 224$ for each sample, which serves as the direct input to our LISA quality prediction model.

\subsection{Illumination-Invariant Quality Prediction: The LISA Framework}
\begin{figure}[t]
  \centering
   \includegraphics[width=1\linewidth]{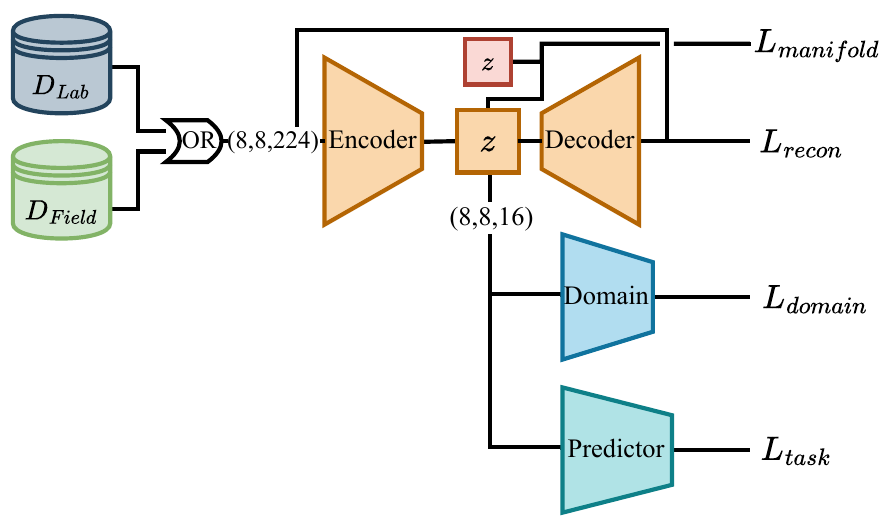}

   \caption{
   \rev{
   Architecture of the proposed Light-Invariant Spectral Autoencoder (LISA). Data from the Lab ($D_{Lab}$) and Field ($D_{Field}$) domains are fed as hyperspectral patches into a central Encoder. The Encoder learns a compressed latent representation ($z$), which serves as a shared feature space for four distinct objectives, each optimized by a corresponding loss function: (1) A Decoder uses $z$ to reconstruct the original input, guided by the reconstruction loss ($L_{recon}$), ensuring features are representative. (2) A Task Predictor regresses grape quality parameters from $z$, guided by the task loss ($L_{task}$). (3) A Domain Discriminator attempts to classify the origin of the data (Lab vs. Field) from $z$. The Encoder is simultaneously trained to ``fool'' this discriminator, an adversarial process governed by the domain loss ($L_{domain}$). This forces $z$ to become domain-invariant. (4) A manifold regularization loss ($L_{manifold}$) is applied to $z$ to promote a semantically meaningful latent space where similar samples are grouped together.
   }}
   \label{fig:modelarcitecture}
\end{figure}

\rev{The core idea of our proposed LISA framework is to learn a feature representation of the hyperspectral data that is highly predictive of grape quality while being invariant to the ambient illumination. Physically, this means the network learns to approximate the grape's intrinsic reflectance by effectively disentangling useful, material-specific spectral features from confounding illumination-related artifacts. This disentanglement is driven by a synergy of learning signals. First, a Domain Discriminator, trained adversarially against the Feature Extractor, provides a strong ``push'' signal, forcing the learned latent features $z$ to become agnostic to their domain of origin. Concurrently, the Task Predictor provides the essential ``pull'' signal, a guidance that compels the Feature Extractor to preserve only the information relevant to the final quality prediction. These competing objectives are refined by a Manifold Regularization loss, which encourages a semantically meaningful latent space by clustering samples with similar Brix values. Our paired dataset design, which isolates illumination as the primary source of variation, further facilitates this learning process. The result is a stable mapping between the latent features and the target quality values, which remains robust even when input spectra vary dramatically due to changing light.}
As illustrated in \cref{fig:modelarcitecture}, the architecture that implements these principles is built on adversarial domain adaptation and consists of three core components:
\begin{itemize}
    \item A Feature Extractor, implemented as a convolutional autoencoder, which learns a compressed latent representation of the spectral data.
    \item A Task Predictor, a Multi-Layer Perceptron (MLP) that regresses the target value (i.e., Brix and Acidity) from the learned features.
    \item A Domain Discriminator, an MLP that attempts to identify the domain origin of the data from the same features.
\end{itemize}
The central component is the Feature Extractor. Its role is to encode the high-dimensional spectral part of the hyperspectral input patch into a low-dimensional latent vector, $z$, with the same spatial dimensions. We implement this using 2D convolutional layers. While hyperspectral data is inherently 3D, we treat the spectral dimension as the channel dimension (i.e., input shape $H\times W \times C$ becomes $8 \times 8 \times224$). This approach effectively leverages a standard 2D CNN architecture to learn spatial-spectral features. The decoder part of the autoencoder then aims to reconstruct the original input from this latent vector $z$.

The learned latent vector $z$ serves as a shared feature representation that is fed into two parallel heads. The Task Predictor uses $z$ to perform the final regression. Simultaneously, the Domain Discriminator also receives $z$ and is trained to classify its source domain (e.g., `Lab' vs. `Field'). During training, the Feature Extractor receives a supervisory signal not only from the reconstruction and prediction losses but also an adversarial signal from the discriminator. This adversarial process, detailed in the following section, forces the extractor to learn a latent space $z$ that is both predictive of the target value and invariant to the input domain.

\subsection{Composite Loss Function}
The LISA framework is trained end-to-end using a composite loss function designed to balance feature reconstruction, task performance, and domain invariance. The total loss, $L_{\text{total}}$, is a weighted sum of four components:
\begin{equation}
    L_{\text{total}} = L_{\text{task}} + \alpha L_{\text{recon}}  + \beta L_{\text{manifold}} + \gamma L_{\text{domain}}
    \label{eq:total_loss}
\end{equation}
where $\alpha$, $\beta$, and $\gamma$ are hyperparameters that balance the contribution of each term. Below, we detail each component.

\begin{itemize}
\item \textbf{Reconstruction Loss ($L_{\text{recon}}$):} This is the standard Mean Squared Error (MSE) between the input spectral patch $X$ and the autoencoder's reconstructed output $\hat{X}$. It ensures the latent space retains sufficient information to represent the original data.
\begin{equation}
    L_{\text{recon}} = ||X - \hat{X}||_F^2
\end{equation}

\item \textbf{Task Prediction Loss ($L_{\text{task}}$):} The primary objective for the regression task is optimized using the MSE between the predicted values $\hat{y}$ and the ground truth labels $y$.
\begin{equation}
    L_{\text{task}} = \frac{1}{N} \sum_{i=1}^{N} (y_i - \hat{y}_i)^2
\end{equation}

\item \textbf{Manifold Regularization ($L_{\text{manifold}}$):} To promote a semantically meaningful latent space, we introduce a manifold regularization term \cite{manifoldREG}. This encourages samples with similar intrinsic properties (i.e., similar Brix values) to be mapped to nearby points in the latent space. For a pair of latent vectors $(z_i, z_j)$, the loss is the squared Euclidean distance weighted by their label similarity:
\begin{equation}
    L_{\text{manifold}} = \sum_{i,j} w_{ij} ||z_i - z_j||_2^2
\end{equation}
where the weight $w_{ij} = \mathbbm{1} $ if the samples belong to the same Brix category, and $w_{ij} = 0$ otherwise.

\item \textbf{Domain Adversarial Loss ($L_{\text{domain}}$):} To achieve domain invariance, we employ an adversarial training strategy inspired by DANN~\cite{DANN}. The training involves a two-player min-max game between the Feature Extractor ($E$) and the Domain Discriminator ($D$). 
\begin{enumerate}
    \item \textbf{Training the Discriminator:} The discriminator $D$ is trained to minimize a standard cross-entropy loss to correctly classify the domain $d$ of the latent features $z = E(X)$.
    \item \textbf{Training the Extractor:} The feature extractor $E$ is trained to \textit{maximize} the discriminator's loss, thereby learning to produce features that are indistinguishable to $D$.
\end{enumerate}
This adversarial objective is implemented using a Gradient Reversal Layer (GRL) \cite{GRL}. The GRL is placed between the encoder and the discriminator. During the forward pass, it acts as an identity function. During backpropagation, it multiplies the gradient from the discriminator by a negative constant ($-1$), effectively reversing its sign. This allows the entire model to be optimized with a single standard backpropagation step, where minimizing $L_{\text{domain}}$ with the GRL in place is equivalent to solving the min-max game.
\end{itemize}

\section{experimental setup}
This section details the experimental protocol used to validate our proposed system. We first outline the evaluation metrics, followed by the specific methodologies for the yield and quality prediction pipelines.

\subsection{Evaluation Metrics}
\subsubsection{Regression}
For the regression tasks of predicting Brix, Acidity, and weight, we use the Coefficient of Determination ($R^2$) as our evaluation metric. $R^2$ measures the proportion of the variance in the dependent variable that is predictable from the model's features. It is defined as:
\begin{equation}
    R^2 = 1 - \frac{\sum_{i=1}^{n} (y_i - \hat{y}_i)^2}{\sum_{i=1}^{n} (y_i - \bar{y})^2}
\end{equation}
where $y_i$ is the ground truth value, $\hat{y}_i$ is the model's prediction, and $\bar{y}$ is the mean of the ground truth values. $R^2$ scores range from $-\infty$ to 1, where 1 indicates a perfect fit, 0 indicates the model is no better than predicting the mean, and negative values indicate a very poor fit. Unlike metrics like MSE, $R^2$ is a normalized score that provides a scale-independent measure of a model's explanatory power, making it ideal for comparing performance across different tasks and datasets. 
\subsubsection{Grape Classification}
For the binary classification task (Grape vs. Non-Grape), we report the Overall Accuracy (OA), defined as the proportion of correctly classified samples.
\subsubsection{Object Detection}
For the object detection task in Pipeline 1, we evaluated the object detection models using standard metrics, including Recall and Mean Average Precision (mAP), averaged over Intersection over Union (IoU) thresholds from 0.5 to 0.95 (mAP@0.5:0.95). For our application, Recall is the most critical metric. A high recall ensures that the system detects the maximum number of grape bunches present in the scene. A failure to detect a bunch (a false negative) means that data is permanently lost for that bunch. In contrast, a lower precision (a false positive) or a slightly imprecise bounding box is less detrimental, as our downstream quality prediction pipeline includes a pixel-level classification step that can effectively filter out any non-grape background included in the detection. Furthermore, this classification step allows us to discard entire bounding boxes if the proportion of grape pixels falls below a set threshold, effectively removing false positive detections.

\subsection{Pipeline 1: Yield Estimation}
The yield estimation pipeline consists of two stages: bunch detection and weight regression.

\subsubsection{Candidate Models and Architectures}
For the bunch detection stage, we evaluated a comprehensive set of state-of-the-art object detectors. This included various sizes from multiple YOLO families (v8\cite{yolo8}, v10\cite{yolo10}, v11\cite{yolo11}, v12\cite{yolo12}) to assess the performance of CNN-based (v8, v10, v11) and attention-centric architectures (v12) of varying sizes. To provide a strong, modern baseline, we also included the Transformer-based RT-DETRv2 model \cite{rtdetrv2}.

To identify the optimal architecture for the bunch weight prediction module, we conducted a comparative evaluation of five candidate models. These models are selected to represent a diverse range of approaches for processing spatio-spectral data: a MLP that treats each spectral band independently; a 1 dimensional convolutional neural network (1D-CNN) that processes the spectrum as a signal \cite{Ref_1D_CNN}; a 2D-CNN that leverages both spatial and spectral context from image patches \cite{Ref_2D_CNN}; a transformer-based model using self-attention to capture global dependencies \cite{Ref_Vision_Transformer}. 

In addition to these standard models, we designed and evaluated a Hybrid CNN-Transformer. This architecture first uses a multi-layer 2D-CNN as a feature extractor to learn a rich set of local spatio-spectral patterns from the input HSI patch. The resulting feature maps are then flattened and treated as a sequence of tokens, which serve as the input to a standard transformer encoder back-end. This two-stage design aims to leverage the efficiency of CNNs for learning local features while using the transformer's self-attention mechanism to model the global relationships between them.

\subsubsection{Evaluation Protocol and Implementation}
For object detection, all YOLO models are implemented using the Ultralytics framework \cite{yolo}. To ensure robust and generalizable results, we evaluated all object detection models and weight prediction models using a k-fold cross-validation scheme on our custom dataset. For the weight prediction models, the input hyperspectral data within each detected bounding box is rescaled to a fixed size of 64x64 pixels before being passed to the model. For each model, we conducted a hyperparameter search to optimize performance, experimenting with key parameters such as learning rate, number of training epochs, and input image resolution. For the object detection model, the final reported models are finetuned from pre-trained COCO weights using the optimized hyperparameters.

\subsection{Pipeline 2: Quality Prediction}
The quality prediction pipeline is designed to predict Brix and Acidity from uncalibrated HSI data while remaining robust to illumination changes.
\subsubsection{Evaluation Scenarios}
\label{sec:scenarios}
To validate the performance of the proposed method, an evaluation is conducted across four distinct scenarios, designed to assess the model's robustness and generalization capabilities under various domain conditions:
\begin{itemize}
    \item Scenario 1 (Intra-Domain): The model is trained and tested exclusively on the Lab dataset, utilizing a standard data split. This scenario establishes a baseline performance under controlled, consistent illumination.
    \item Scenario 2 (Cross-Domain: Lab to Field): The model is trained on the Lab dataset and subsequently tested on the Field-AM and Field-PM datasets separately. This scenario directly evaluates the model's transferability and robustness to the significant domain shift encountered when moving from controlled laboratory conditions to dynamic field environments.
    \item Scenario 3 (Cross-Domain: Field to Field): The model is trained on the Field-AM dataset and then tested on the Field-PM dataset. This scenario assesses the model's ability to generalize between different natural illumination conditions within the field, specifically accounting for the natural shift in solar angle and intensity between morning and afternoon.
    \item Scenario 4 (Domain Generalization): The model is trained on a combined dataset of Lab and Field-AM data, and then tested on the Field-PM dataset or vice versa. This scenario investigates the benefits of training with a more diverse set of source domains to improve generalization to a target field domain.
\end{itemize}

\subsubsection{Baseline Methods for Comparison}
\label{sec:baselines}
\begin{itemize}
    \item \textbf{Partial Least Squares (PLS) \cite{hsigrape0, hsigrape2}:} A standard linear chemometric technique that is highly effective for handling high-dimensional and collinear spectral data. It works by projecting the spectral variables and the response variables to a new space of latent variables, optimizing for covariance.
    \item \textbf{Predictor-Only Baseline:} This serves as a fundamental baseline, evaluating the performance of only the MLP predictor without the benefit of the autoencoder's learned feature space. 
    \item \textbf{LOGSEP \cite{LOGSEP} Method:} We implemented the LOGSEP method for illumination correction as a classic physics-based baseline. This approach aims to derive reflectance spectra independent of illumination. The resulting reflectance spectra are then fed into the same MLP predictor as our proposed model, ensuring a fair comparison of the illumination correction technique itself.
    \item \textbf{SILL-R-GAN \cite{SILLRGAN}:} We compared our approach against SILL-R-GAN, a state-of-the-art GAN-based method specifically designed for illumination-robust HSI classification. To enable a direct comparison, its generator component is adapted to produce illumination-invariant spectra. These spectra then served as input to the model's original classification head, which is repurposed for the regression task.
\end{itemize}

\subsubsection{Implementation and Hyperparameter Tuning}
\label{hyperparameter}
All deep learning models for pipeline 2, including our proposed LISA framework and the baselines, are implemented in PyTorch. To ensure a fair and robust comparison, each model is optimized via an extensive randomized hyperparameter search managed with the Weights \& Biases platform \cite{wandb}. This search encompassed model-specific architectural choices, training dynamics, and data preprocessing pipelines.

A key finding from this process is that a Savitzky-Golay first-derivative filter consistently yielded the best performance for all models, including the baselines. Therefore, this preprocessing step is adopted for all quality prediction experiments reported in this paper. The remainder of the hyperparameter search then focused on optimizing the unique architectural and training parameters for each model.

For each domain generalization scenario presented in \cref{sec:results}, the optimal hyperparameters for each model are determined independently. The final configuration detailed below corresponds to LISA's best-performing model on the primary task of generalizing from Lab+PM to the Field-AM target domain. This model is trained for 200 epochs using the Adam optimizer with a learning rate of $1.5 \times 10^{-4}$ for the autoencoder and $2.5 \times 10^{-4}$ for the domain discriminator. The key loss weights are set to $\alpha=0.011$ (reconstruction), $\beta=0.066$ (manifold), and $\gamma = 1.2 \times 10^{-4}$ (adversarial).

\section{Results}
\label{sec:results}
This section evaluates the performance of our proposed system by systematically validating each of its core components. First, we establish the performance of the yield estimation pipeline, identifying the optimal models for bunch detection and weight prediction. Second, we evaluate the illumination-invariant quality pipeline, demonstrating how our proposed LISA framework addresses the challenge of domain shift. Finally, we showcase the end-to-end system's capability by demonstrating its integrated output for generating in-field quality and yield data.

\subsection{Pipeline 1: Yield Estimation Performance}
Accurate yield estimation requires the precise detection and localization of individual grape bunches within the hyperspectral image or a corresponding RGB image. To address this, we evaluated a suite of state-of-the-art object detection models from Ultralytics \cite{yolo}, the YOLO (You Only Look Once) family. We focused on finding a model that provides the best trade-off between detection accuracy and recall.

\subsubsection{Bunch Detection (YOLO)}

\begin{table}[!t]
    \centering
    % Refined caption to include the context of model parameters
    \caption{Cross-Validated Performance of Object Detection Models. The table includes parameter counts to illustrate the trade-off between model size and performance. Performance is judged primarily on Recall, with the strict mAP@0.5:0.95 as a secondary metric. The \textbf{YOLOv11-Large} model is selected for its balance of high recall and model efficiency.}
    \vspace{2mm}
    \label{tab:yolo_comparison_final}
    % Added a new 'c' column for Parameters
    \begin{tabular}{@{}l l c c c@{}}
        \toprule
        \textbf{Model Family} & \textbf{Size} & \textbf{Params (M)} & \textbf{Recall} & \textbf{mAP@0.5:0.95} \\
        \midrule
        % --- YOLOv8 Block ---
        \multirow{5}{*}{YOLOv8 \cite{yolo8}} 
        & Nano (n) & 3.2 & 0.73 & 0.57 \\
        & Small (s) & 11.2 & 0.72 & 0.55 \\
        & Medium (m) & 25.9 & 0.80 & 0.62 \\
        & Large (l) & 43.7 & 0.78 & 0.64 \\
        & X-Large (x) & 68.2 & 0.71 & 0.46 \\
        \cmidrule(l){2-5} % Extended rule to cover the new column
        
        % --- YOLOv10 Block ---
        \multirow{5}{*}{YOLOv10 \cite{yolo10}} 
        & Nano (n) & 2.3 & 0.70 & 0.52 \\
        & Small (s) & 7.2 & 0.79 & 0.64 \\
        & Medium (m) & 15.4 & 0.80 & 0.62 \\
        & Large (l) & 24.4 & 0.79 & 0.65 \\
        & X-Large (x) & 29.5 & 0.78 & 0.66 \\
        \cmidrule(l){2-5}
        
        % --- YOLOv11 Block (Winning Model) ---
        \multirow{5}{*}{YOLOv11 \cite{yolo11}} 
        & Nano (n) & 2.6 & 0.76 & 0.59 \\
        & Small (s) & 9.4 & 0.78 & 0.63 \\
        & Medium (m) & 20.1 & 0.81 & 0.65 \\
        & \textbf{Large (l)} & \textbf{25.3} & \textbf{0.82} & \textbf{0.66} \\
        & X-Large (x) & 56.9 & 0.76 & 0.64 \\
        \cmidrule(l){2-5}
        
        % --- YOLOv12 Block ---
        \multirow{5}{*}{YOLOv12 \cite{yolo12}} 
        & Nano (n)  & 2.6 & 0.80 & 0.63 \\                
        & Small (s) & 9.3 & 0.80 & 0.65 \\
        & Medium (m) & 20.2 & 0.80 & 0.63 \\
        & Large (l) & 26.4 & 0.80 & 0.64 \\
        & X-Large (x) & 59.1 & 0.78 & 0.62 \\    
        \cmidrule(l){2-5}
        
        % --- RT-DETR Block ---
        \multirow{2}{*}{RT-DETRv2 \cite{rtdetrv2}} 
        & Large (l) & 42.0 & 0.82 & 0.64 \\
        & X-Large (x) & 76.0 & 0.73 & 0.60 \\
        \bottomrule
    \end{tabular}
\end{table}

To select the optimal bunch detector for our pipeline, we conducted a comprehensive evaluation of state-of-the-art object detection models. This included various sizes from multiple YOLO families (v8, v10, v11, and v12) as well as the transformer-based RT-DETRv2 \cite{rtdetrv2}, which serves as a modern baseline comparable to the architecture used in the work by Bertoglio et al.  \cite{hsigrape2}. Performance is judged primarily on Recall, given its importance for minimizing data loss, with mAP@0.5:0.95 as a secondary metric for localization accuracy. The cross-validated results are summarized in \cref{tab:yolo_comparison_final}.

The results highlight several key trends. Across all families, there is a general, though not universal, trend of diminishing returns with the largest model sizes (X-Large). The recently introduced YOLOv12 family demonstrates remarkably consistent high performance, with most variants achieving a Recall of 0.80. Similarly, the RT-DETRv2 baseline performs competently, achieving a recall of 0.80 with its large variant, confirming its suitability for this task.

Despite the strong performance from these newer models, the YOLOv11-Large (l) model remains the superior choice for our application. It achieved the highest Recall of all tested models at 0.82, ensuring the maximum number of grape bunches are detected. Furthermore, its mAP@0.5:0.95 score of 0.66 is among the best, matching the performance of the YOLOv10-X-Large model. This accuracy is achieved with a significantly more efficient architecture, having 17\% fewer parameters and requiring 50\% fewer FLOPs than the comparable X-Large variant \cite{yolo}, making it a better choice for robotic deployment. Based on this combination of recall and high computational efficiency, we selected the YOLOv11-Large model to drive the yield estimation pipeline.

\subsubsection{Bunch Weight Prediction}
Once a grape bunch is accurately localized by the YOLOv11-l detector, the second stage of the yield estimation pipeline predicts its weight from the enclosed hyperspectral data. To identify the most effective architecture for this regression task, we evaluated several candidate architectures, including an MLP, 1D/2D-CNNs, a transformer, and a hybrid model. The performance of these models, measured by the $R^2$, is presented in \cref{tab:weight_prediction}.

\begin{table}[h!]
\centering
\caption{Cross-Validated Performance Comparison for Bunch Weight Prediction (g). The best performance is shown in bold.}
\vspace{2mm}
\label{tab:weight_prediction}
\begin{tabular}{l c}
\toprule
\textbf{Model Architecture} & \textbf{$R^2$} \\
\midrule
MLP                        & 0.59 \\
1D CNN                         & 0.57 \\
2D CNN                      & \textbf{0.76} \\
Transformer                         & 0.50 \\
% \multicolumn{4}{l}{\textit{Deep Learning Methods}} \\
CNN + Transformer           & 0.34 \\
% Add other DL models if you want, but CNN+Transformer is enough to make the point
\bottomrule
\end{tabular}
\end{table}

The superior performance of the 2D CNN architecture ($R^2$=0.76) suggests that leveraging local spatio-spectral information is critical for accurate weight prediction. Standard MLP and 1D-CNN models performed moderately but are limited by their inability to process spatial context. Similarly, the transformer-only model struggled, indicating that for this task, local textural features are more informative than global spectral relationships.

Notably, the hybrid CNN-Transformer, which theoretically combines the strengths of both architectures, performed the worst. We hypothesize this is due to its significantly higher model complexity. Such a high-capacity model is prone to overfitting, failing to generalize beyond the training set. In contrast, the simpler 2D-CNN architecture strikes the right balance, effectively learning predictive local features likely corresponding to physical characteristics like berry density and internal shadowing without overfitting. Therefore, we selected the 2D CNN as the optimal regression model for our yield estimation pipeline.

\subsection{Pipeline 2: Illumination-Invariant Quality Prediction (LISA)}
The second core component of our integrated system is the quality prediction pipeline, which assesses intrinsic grape properties like Brix and Acidity from HSI data. The primary challenge for this task in field robotics is the significant data variability caused by changing illumination. A model that is not robust to this ``domain shift" will fail when deployed in real-world conditions. 

This section details the evaluation of our proposed solution, the Light-Invariant Spectral Autoencoder. We first quantitatively demonstrate the severity of the domain shift problem using a baseline model. We then present LISA's main results, showing its superior generalization performance compared to state-of-the-art methods. Finally, we provide a series of analyses and ablation studies to validate our model's design and understand the sources of its robustness.
\subsubsection{Quantifying Domain Shift}
The Predictor-Only baseline is trained and tested on all nine possible combinations of our three domains (Lab, Field-AM, Field-PM), i.e., the first 3 scenarios explained in \cref{sec:scenarios}. The results, measured by the $R^2$, are presented in \cref{tab:cross_domain_mlp}.

\begin{table}[t!]
\centering
\caption{Cross-Domain and Within-Domain Brix Prediction Performance ($R^2$) for the Predictor-Only baseline. Bold values on the diagonal represent within-domain performance, highlighting the performance degradation in cross-domain scenarios.}
\vspace{2mm}
\label{tab:cross_domain_mlp}
\begin{tabular}{l ccc}
\toprule
% Header Row 1: Leave the first column empty
& \multicolumn{3}{c}{\textbf{Test Domain}} \\
\cmidrule(lr){2-4}
% Header Row 2: Full headers
\textbf{Training Domain} & \textbf{Lab} & \textbf{Field-AM} & \textbf{Field-PM} \\
\midrule
Lab         & \textbf{0.68} & 0.29          & 0.15 \\
Field-AM    & 0.19          & \textbf{0.73} & 0.12 \\
Field-PM    & 0.16          & 0.15          & \textbf{0.54} \\
\bottomrule
\end{tabular}
\end{table}

The results in \cref{tab:cross_domain_mlp} clearly illustrate a drastic performance degradation when the model is applied across domains. The diagonal values, representing intra-domain performance, show that the Predictor-Only baseline can achieve reasonable results when trained and tested on data from the same environment, with $R^2$ scores of 0.68 (Lab), 0.73 (Field-AM), and 0.54 (Field-PM). Notably, the performance on the Field-PM subset is inherently lower, which we hypothesize is due to reduced image sharpness from slight changes in camera distance in the field, an effect that was not compensated for during measurements.

However, in all off-diagonal, cross-domain scenarios, the model's predictive power collapses. For instance, a model trained in the controlled Lab environment ($R^2$=0.68) fails to generalize to either of the field settings, with its performance plummeting to an $R^2$ of 0.29 on Field-AM data and just 0.15 on Field-PM data. A similar drop is observed when training in the field and testing in the lab. This severe and consistent performance degradation provides a quantitative confirmation of the substantial domain gap between the datasets.

\subsubsection{Generalization Performance}
Having established the challenge of domain shift, we now evaluate LISA's generalization performance using a ``leave-one-domain-out" protocol. 
\rev{The dataset is composed of three distinct acquisition domains. The shift from controlled artificial light (Lab) to the continuously evolving natural sunlight in the morning and afternoon represents a set of complex, non-linear changes in illumination intensity and spectral composition (as shown in \cref{fig:short-b}). LISA's ability to consistently outperform all baselines in these challenging leave-one-domain-out scenarios serves as robust quantitative evidence of its effectiveness under diverse and challenging field conditions.}
The different models are trained on a combination of two domains and tested on the held-out third, with results against baselines presented in \cref{tab:domain_adaptation}.
\begin{table*}[t!]
\centering
\caption{Domain Generalization Performance on Unseen Target Domains. Models are trained on a combination of two domains and tested on the held-out third. Performance is measured by $R^2$ for regression tasks (higher is better, ↑) and overal accuracy (OA) for classification (higher is better, ↑).The best performance for each metric is shown in bold.}
\vspace{2mm}
\label{tab:domain_adaptation}
\begin{tabular}{l ccc ccc}
\toprule
& \multicolumn{3}{c}{\textbf{Target: Field-AM} (Train on Lab+PM)} & \multicolumn{3}{c}{\textbf{Target: Field-PM} (Train on Lab+AM)} \\
\cmidrule(lr){2-4} \cmidrule(lr){5-7}
\textbf{Model}      & \textbf{Brix ($R^2$) ↑} & \textbf{Acid ($R^2$) ↑} & \textbf{Grape (OA) ↑} & \textbf{Brix ($R^2$) ↑} & \textbf{Acid ($R^2$) ↑} & \textbf{Grape (OA) ↑} \\
\midrule
Predictor-Only   & 0.41                  & 0.37                  & 0.97                  & 0.19                  & 0.21                  & 0.95 \\
PLS \cite{hsigrape0, hsigrape2} & 0.25 & 0.37 & 0.98 & 0.13 & 0.18 & 0.97 \\
LOGSEP \cite{LOGSEP}        & 0.17                  & 0.07                  & \textbf{1.00}                  & 0.33                  & 0.02                  & \textbf{1.00} \\
SiLLR-GAN \cite{SILLRGAN}     & 0.43                  & 0.43                  & \textbf{1.00}                  & 0.38                  & 0.29                  & \textbf{1.00} \\
\textbf{LISA} (ours)  & \textbf{0.62}         & \textbf{0.47}         & \textbf{1.00}         & \textbf{0.40}         & \textbf{0.38}         & 0.99 \\
\bottomrule
\end{tabular}
\end{table*}

In the regression tasks, our model consistently outperforms all baselines. When generalizing to the Field-AM target domain, LISA achieves an $R^2$ of 0.62 for Brix prediction. This is a substantial improvement of 44\% over the next-best deep learning method, SiLLR-GAN ($R^2 = 0.43$), and demonstrates clear superiority over the Predictor-Only baseline ($R^2 = 0.41$). 
It also significantly outperforms classical methods, including Partial Least Squares regression ($R^2 = 0.25$).
\rev{
A crucial result from this analysis is the direct comparison between our data-driven LISA framework and the physics-based LOGSEP baseline. The severe underperformance of LOGSEP (e.g., $R^2$ of 0.17 when generalizing to Field-AM) underscores the limitations of methods based on simplified physical assumptions. Real-world vineyard illumination is a composite of direct sunlight, atmospheric scattering, and complex, non-linear reflections and shadows from the canopy and within the bunch itself, which are not adequately captured by linear models. In contrast, LISA's ability to learn these complex relationships directly from data allows it to achieve significantly more robust and accurate predictions, demonstrating the practical superiority of a data-driven approach for this application.}
A similar trend is observed when generalizing to the more challenging Field-PM target, where our model leads with an $R^2$ of 0.40. The performance on Acidity prediction follows the same pattern, with our model achieving the highest $R^2$ in both scenarios (0.47 and 0.38, respectively). This superior performance indicates that our adversarial autoencoder learns a more robust feature representation that is less sensitive to illumination changes, making it better suited for predicting intrinsic chemical properties.

A key advantage of our HSI-based quality pipeline is its ability to perform implicit pixel-level masking. Unlike methods that rely on pre-segmenting grapes from RGB images \cite{hsigrape2}, which can be imprecise and lead to contamination from stems or background, LISA's feature extractor learns to distinguish grape from non-grape pixels directly from the rich spectral data. This ensures that only pure grape spectra contribute to the final regression.

\subsubsection{Analysis and Ablation Studies for LISA}

To better understand the sources of LISA's robustness, we conducted a series of analytical experiments and ablation studies.

First, to validate our architectural choices, we evaluated several design variants for the feature extractor, with results shown in \cref{tab:arch_choices}. The standard 2D CNN autoencoder provided the best performance ($R^2$=0.62) on the domain generalization task. While a 3D CNN is theoretically a good fit for HSI data, its higher parameter count makes it significantly more prone to overfitting the training data. Its lower performance suggests that it learned complex spatio-spectral features that did not generalize well to the unseen test set. This result confirms that our more streamlined 2D CNN design strikes a better balance between model capacity and generalization, making it the most effective core for LISA.

Second, we performed an ablation study on the composite loss function to isolate the contribution of each component (\cref{tab:ablation}). The results demonstrate the synergistic importance of our complete loss design. The most critical component is clearly the adversarial loss ($L_{\text{domain}}$); its removal causes the largest performance degradation ($R^2$ drops to 0.50), confirming that forcing domain-invariance is the primary driver of generalization. The manifold regularization and reconstruction losses also provide positive contributions, guiding the model to a more meaningful solution and preventing feature collapse.

\begin{figure}[ht]
    \centering
    \includegraphics[width=\columnwidth]{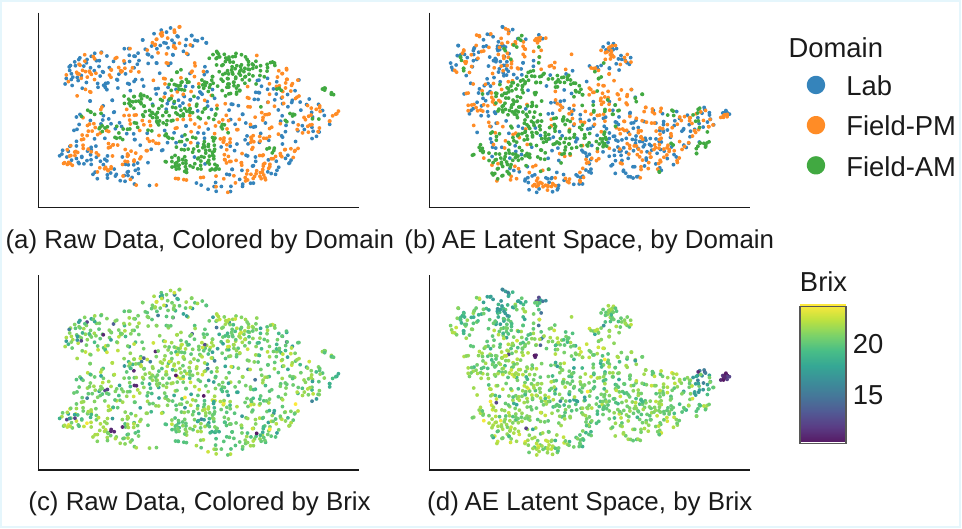}
    \caption{
    \rev{
    t-SNE visualization demonstrating the effectiveness of LISA in learning domain-invariant and task-relevant features.
(Top Row: Domain Invariance) Points are colored by their acquisition domain (Lab, Field-AM, Field-PM). While the raw data (a) shows more distinct clustering based on lighting conditions, the learned latent space (b) shows that the three domains are more intermixed, confirming the model has learned features that are invariant to the domain shift.
(Bottom Row: Task-Relevant Structure) Points are colored by their ground truth Brix value. The model transforms the unstructured distribution in the raw data (c) into a more continuous manifold in the latent space (d), where samples with similar Brix values are located near each other. This structured representation is ideal for robust regression.
}
}
    \label{fig:latent space}
\end{figure}

\rev{Finally, to assess the model's behavior, we visualized the learned latent space using t-SNE (\cref{fig:latent space}). The top row shows that while raw data points cluster by domain (\cref{fig:latent space}a), the latent features from LISA are more mixed (\cref{fig:latent space}b). This visual observation is confirmed quantitatively: the accuracy of a support vector machine (SVM) \cite{svm} classifier for domain prediction dropped by 15\% (from 75.6\% to 64.6\%), and the Maximum Mean Discrepancy (MMD) \cite{MMD} between domain distributions decreased by 27\%. While greater domain confusion could be achieved by increasing the weight of the adversarial loss, our configuration strikes a deliberate balance between achieving domain invariance and maintaining high performance on the primary regression task. This process ultimately organizes the unstructured raw data (\cref{fig:latent space}c) into a more coherent manifold based on the target Brix value (\cref{fig:latent space}d), which is an ideal feature representation for robust regression.}

\begin{table}[h!]
\centering
\caption{Comparison of different architectural choices for the feature extractor on the Field-AM subset. The 2D CNN autoencoder is selected for the final LISA model.}
\vspace{2mm}
\label{tab:arch_choices}
\begin{tabular}{l c}
\toprule
\textbf{Architectural Variant} & \textbf{Brix ($R^2$)}  \\
\midrule
\textbf{2D CNN Autoencoder (Chosen)} & \textbf{0.62} \\
3D CNN Autoencoder             & 0.51 \\
Variational AE (2D CNN)        & 0.60 \\
Bi-Classifier Architecture     & 0.60 \\
\bottomrule
\end{tabular}
\end{table}

\begin{table}[h!]
\centering
\caption{Ablation study of LISA's loss components. The full model is compared against variants with a single loss term removed on the Field-AM subset.}
\vspace{2mm}
\label{tab:ablation}
\begin{tabular}{l c}
\toprule
\textbf{Model Variant} & \textbf{Brix ($R^2$)}  \\
\midrule
\textbf{LISA (Full Model)} & \textbf{0.62} \\
\midrule
\textit{Ablations:} \\
\quad - w/o Adversarial Loss ($\mathcal{L}_{\text{domain}}$)  & 0.50 \\
\quad - w/o Manifold Reg. ($\mathcal{L}_{\text{manifold}}$) & 0.56 \\
\quad - w/o Reconstruction Loss ($\mathcal{L}_{\text{recon}}$) & 0.60 \\
\bottomrule
\end{tabular}
\end{table}

\subsection{Integrated In-Field Yield and Quality Mapping}
The ultimate goal of our robotic system is not to perform individual predictions, but to synthesize this information into a spatially-aware map that can guide agricultural decisions. This is accomplished by integrating the optimal components identified in our previous analyses: the \textbf{YOLOv11-l} detector, selected for its Recall (0.82); the \textbf{2D CNN} weight regressor, chosen for its superior $R^2$ of 0.76; and our novel \textbf{LISA} framework for robust quality assessment across domains. \rev{Crucially, this entire analytical pipeline is efficient, capable of processing a full vineyard row scan on a CPU (Intel Core i5-10210U) without GPU in approximately one-third of the time required for data acquisition, enabling real-time operation.}

\rev{
To quantitatively validate the end-to-end reliability of the integrated pipeline, which combines the optimal yield, quality, and detection models identified in our component-wise evaluations. We performed a final evaluation on a completely held-out test scan containing 13 grape bunches, captured in the field under lighting conditions intermediate to the main AM and PM training domains. Given the small sample size of this scan (N=13), we report performance using Mean Absolute Error (MAE) and Mean Absolute Percentage Error (MAPE). For the primary quality task, the system achieved a MAE of 0.92° Brix (4.24\% MAPE). For the other parameters, it achieved a MAE of 0.12 g/L for acidity (15.8\% MAPE) and 44.8 g for weight (32.2\% MAPE). These results from a novel in-field scan confirm the system's practical performance and its potential for reliable, on-the-fly mapping.}

\begin{figure}[t]
  \centering
   \includegraphics[width=0.9\linewidth]{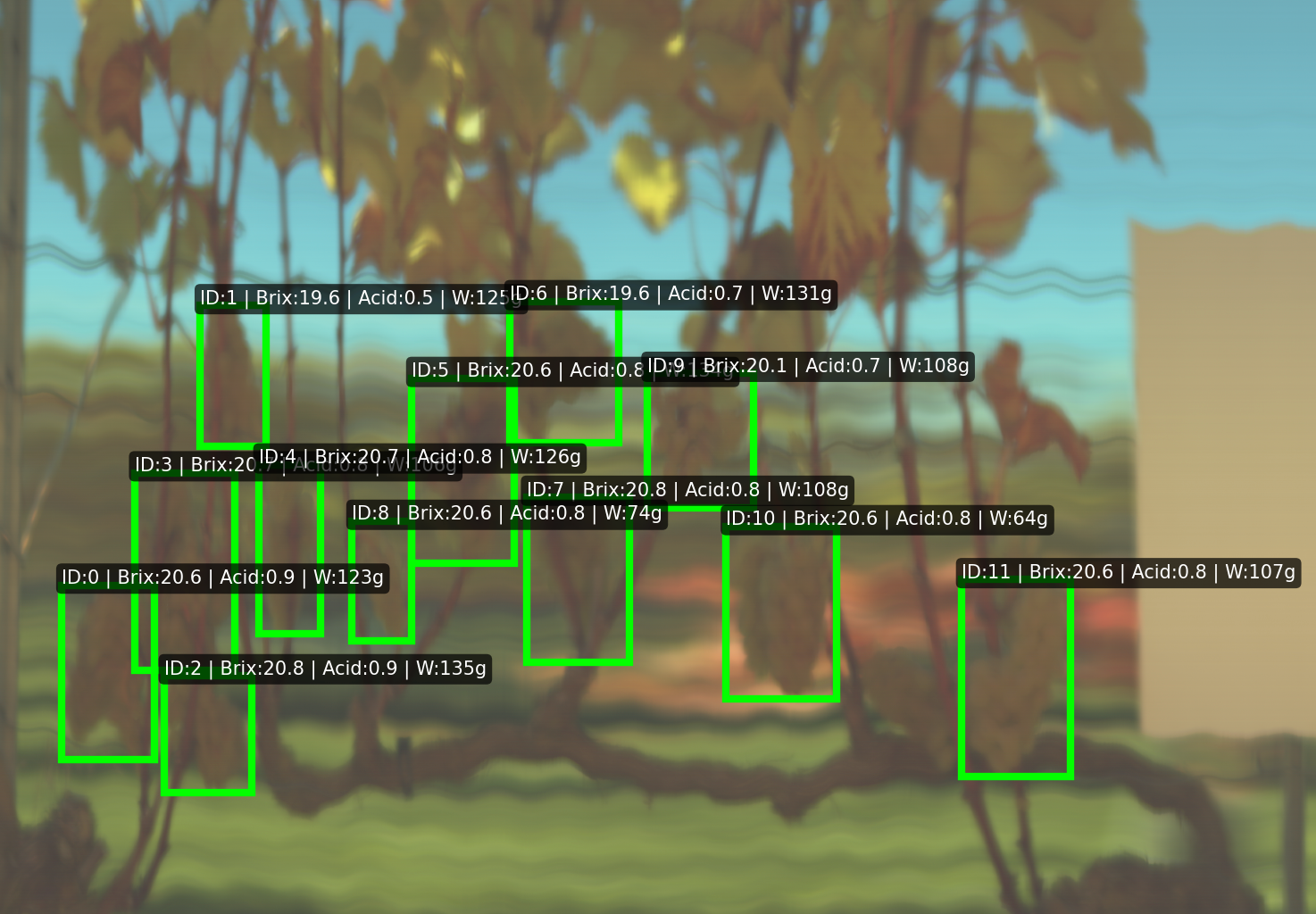}

   \caption{Visualization of the integrated system's output on a single hyperspectral scan captured by the robotic platform. The YOLO-based module detects individual grape bunches (green bounding boxes). For each detected bunch, the system predicts and displays its unique identifier (ID), estimated weight (W), and the average predicted quality parameters (Brix and Acidity) derived from the LISA framework. This demonstrates the system's ability to generate rich, spatially-resolved data points in a complex, in-field environment, which can then be used to create comprehensive vineyard maps.}
   \label{fig:integrated_output}
\end{figure}

\Cref{fig:integrated_output} showcases the per-bunch data generated by the system from a small window of a single hyperspectral scan captured in the field. This visualization serves as a demonstration of the end-to-end pipeline in action. Each detected bunch is successfully localized and annotated with its unique tracker ID, its predicted weight (W), and the average Brix and Acidity values derived from the LISA model.
Notably, the predictions in the figure reveal significant intra-vineyard variability, even within this single field of view. For instance, predicted Brix values range from 19.6 to 20.8, while estimated weights vary dramatically from 64g to 135g. This is precisely the granular, high-frequency data that traditional destructive sampling methods fail to capture.

Each of these predictions is associated with the GPS coordinates captured at the time of the scan. By aggregating these individual georeferenced data points as the robot traverses the vineyard, a high-resolution map of quality and yield can be generated. These maps allow growers to visualize trends, identify underperforming zones, and delineate management areas for targeted interventions, such as selective harvesting or variable-rate fertilization, thereby fulfilling the core promise of precision viticulture.

\section{conclusion}
The practical deployment of autonomous robotic systems for precision agriculture is often impeded by environmental variability. In this work, we addressed the critical challenge of changing illumination for hyperspectral sensing by developing an end-to-end, IoT-enabled robotic system for in-field grape quality and yield assessment. The core of our contribution is the Light-Invariant Spectral Autoencoder, a novel domain-adversarial framework that learns robust features directly from uncalibrated hyperspectral data.

Through experiments on a unique multi-domain dataset, LISA demonstrated superior generalization, outperforming baseline methods by over 20\% in predicting Brix and Acidity under unseen lighting conditions. Ablation studies confirmed that the adversarial component is the key driver of this robustness. We integrated this novel quality prediction module with a high-performing yield estimation pipeline, for which we identified a YOLOv11-Large for bunch detection (0.82 Recall) and a 2D CNN for weight prediction (0.76 $R^2$) as the optimal architectures.

This work represents a significant step towards developing truly autonomous and field-deployable HSI systems. By tackling the fundamental challenge of illumination invariance at the algorithmic level, our integrated system reduces the need for impractical field calibrations and successfully generates the high-resolution, georeferenced data of both quality and yield required to enable data-driven decision-making in precision viticulture.

\section{Limitations and Future Work}
While this work demonstrates a significant step forward, we acknowledge several limitations that open avenues for future research.
\begin{itemize}
    \item Bridging the Performance Gap: While LISA marks a substantial improvement over baselines, the absolute performance for cross-domain quality prediction (i.e., $R^2$ of 0.62 for Brix) indicates a remaining gap to the level of accuracy required for high-stakes commercial decisions. Future research should focus on bridging this gap.
    \rev{
    \item Data-driven Approach: The effectiveness of our data-driven approach is contingent on the training dataset's diversity. While our current dataset was acquired during a focused one-day collection campaign, a key limitation is the substantial upfront investment required to scale this effort. Capturing a wider range of grape varieties, growth stages, and environmental conditions would necessitate a significantly larger data collection campaign. Future work could explore methods like advanced data augmentation or semi-supervised learning to mitigate this data acquisition burden.
    }
    \rev{\item Dataset Scope: Our validation is performed on a single grape variety (Chardonnay) and location. Future work should test the system's robustness across different varieties, which possess unique spectral signatures \cite{red_wine}, as well as across different growth stages and geographical locations to further validate its generalization capabilities.} Investigating performance on fully unpaired datasets would also be a valuable next step.
    \rev{\item
    Stateless vs. Temporal Modeling: LISA’s stateless, per-image design ensures operational flexibility and robustness to abrupt illumination shifts (e.g., moving between sun and shadow). However, it does not leverage temporal correlations in changing sunlight. Future work could explore hybrid architectures that fuse temporal and instantaneous features. For example, a recurrent layer could process a sequence of recent spectral features to model the gradual drift in illumination. This could enhance stability while retaining the benefits of a stateless approach.
    }
    \item Physical Model Integration: LISA operates as a data-driven model. Integrating explicit physical light transport principles or atmospheric correction models into the adversarial framework could further improve performance and yield more interpretable, calibrated reflectance features.
    \item Advanced Model Architectures: While our 2D-CNN backbone for LISA proved effective, exploring more sophisticated architectures, such as Vision Transformers (ViTs) for either the feature extractor or the prediction heads, could potentially unlock further performance gains from the rich spatio-spectral data.
    \item Expansion of Prediction Tasks: The framework can be extended to predict other critical quality parameters, such as pH, tannins, or flavonols, which are also usefull for winemaking decisions \cite{extraChemicals}.
\end{itemize}
Addressing these areas will continue to bridge the gap between promising research and the widespread deployment of intelligent robotic systems in real-world agricultural environments.

\section*{Data Availability Statement}
The dataset generated and analyzed for this study is available from the corresponding author upon reasonable request. A public release of the dataset is planned for the future.

\section*{Acknowledgments}
This work received financial support from the Flanders AI Research Program (FAIR). We also extend our gratitude to the research team at the Institute for Agriculture, Fisheries and Food Research (ILVO) for their crucial support and expertise during the in-field data acquisition campaigns and for the subsequent chemical analysis of the grape samples. We are also grateful to Vicky Everaerts of pcfruit for providing access to the vineyard, and to Aaron Van Gehuchten of ILVO for his assistance with the in-field data acquisition. Sander De Coninck receives funding from the Special Research Fund of Ghent University under grant no. BOF22/DOC/093

\bibliographystyle{ieeetr}
\bibliography{main}
\vspace{-10mm} % Adjust this value as needed

\begin{IEEEbiography}[{\includegraphics[width=1in,height=1.25in,clip,keepaspectratio]{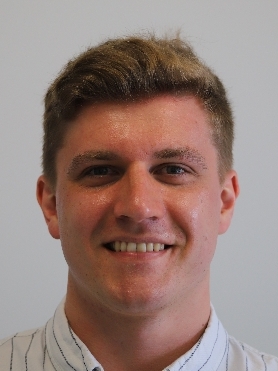}}]{Ciem Cornelissen}is currently pursuing his PhD in Computer Science from Ghent University, Belgium since June 2024, and is also affiliated with imec. 

His main research interests are deep learning and computer vision for robotic sensing.
\end{IEEEbiography}
\vspace{-10mm} % Adjust this value as needed

\begin{IEEEbiography}[{\includegraphics[width=1in,height=1.25in,clip,keepaspectratio]{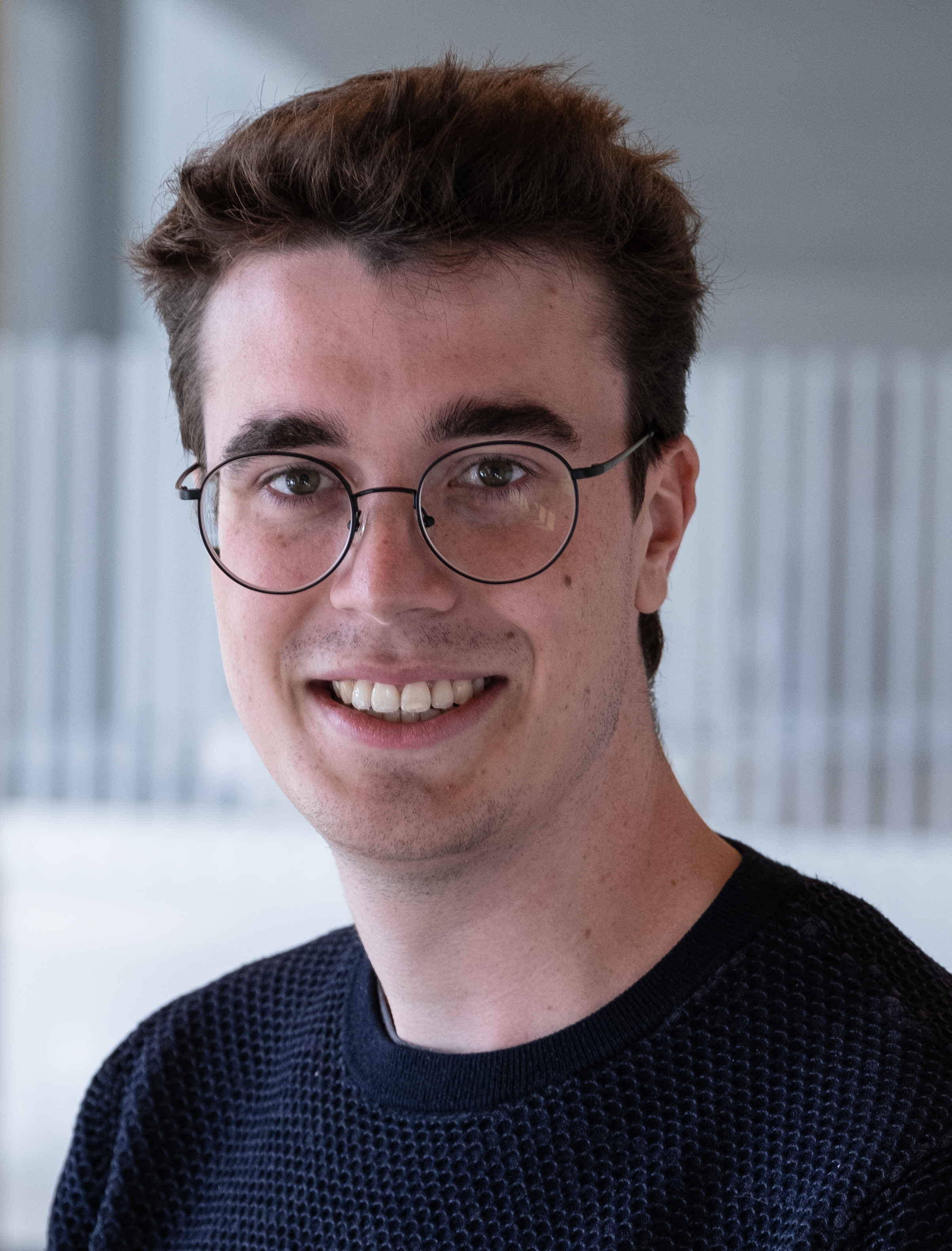}}]{Sander De Coninck}is currently pursuing his PhD in Information Engineering Technology from Ghent University, Belgium since September 2021, and is also affiliated with imec. 

His main research interests are privacy-preserving and fair machine learning and computer vision.
\end{IEEEbiography}
\vspace{-10mm} % Adjust this value as needed

\begin{IEEEbiography}[{\includegraphics[width=1in,height=1.25in,clip,keepaspectratio]{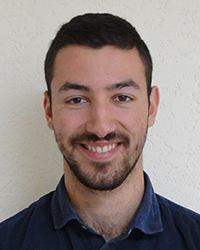}}]{Axel Willekens} is a PhD researcher at Flanders Research Institute for Agriculture, Fisheries and Food (ILVO) Merelbeke-Melle, Belgium.

His research focuses on the development and integration of task and navigation algorithms for autonomous precision crop farming applications.
 
\end{IEEEbiography}
\vspace{-10mm} % Adjust this value as needed

\begin{IEEEbiography}
[{\includegraphics[width=1in,height=1.25in,clip,keepaspectratio]{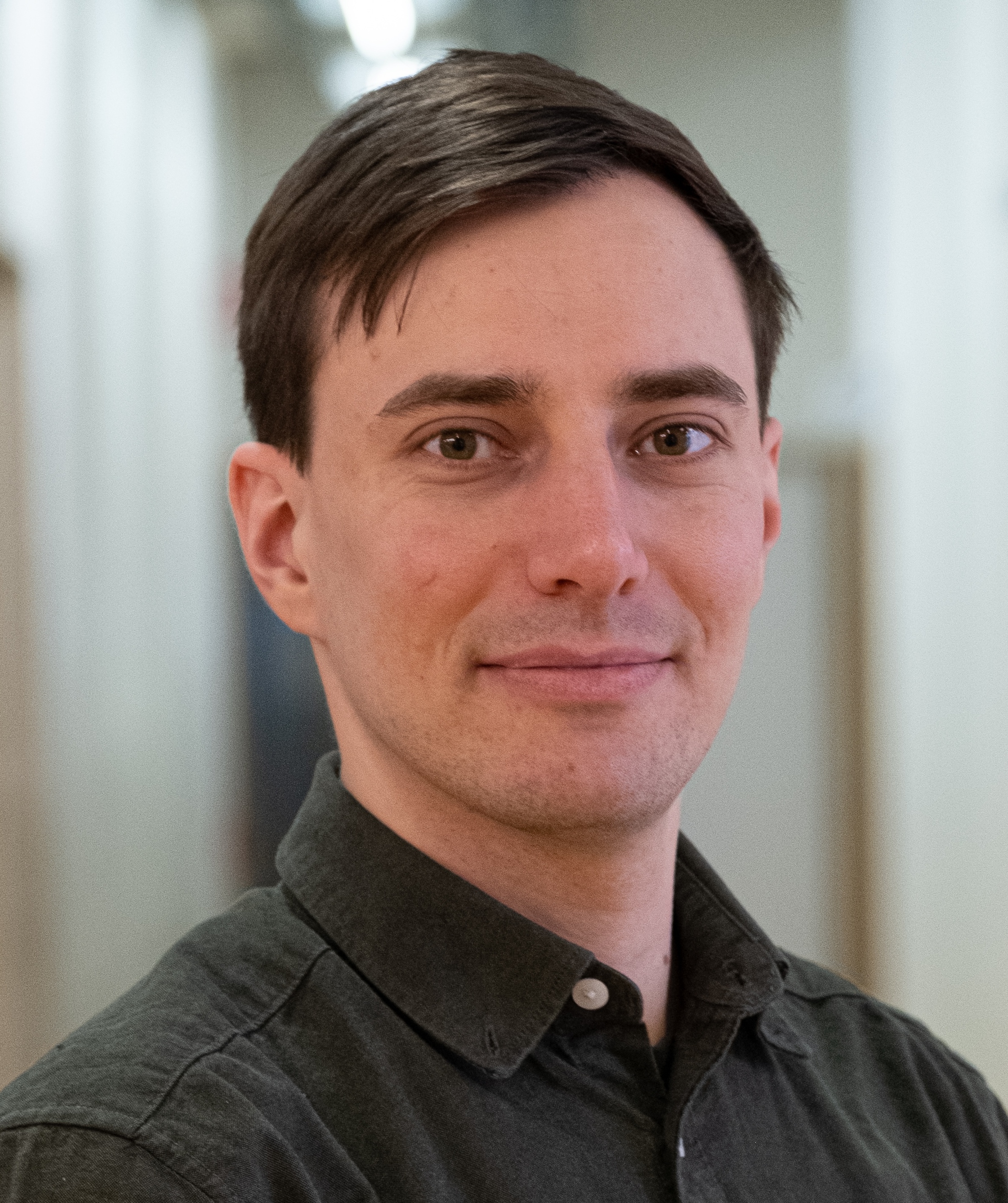}}]{Sam Leroux}
is an assistant professor at the IDLab, an IMEC research group at Ghent University, Ghent, Belgium. 

His main research interests include efficient neural network architectures and edge computing. 
\end{IEEEbiography}
\vspace{-10mm} % Adjust this value as needed

\begin{IEEEbiography}
[{\includegraphics[width=1in,height=1.25in,clip,keepaspectratio]{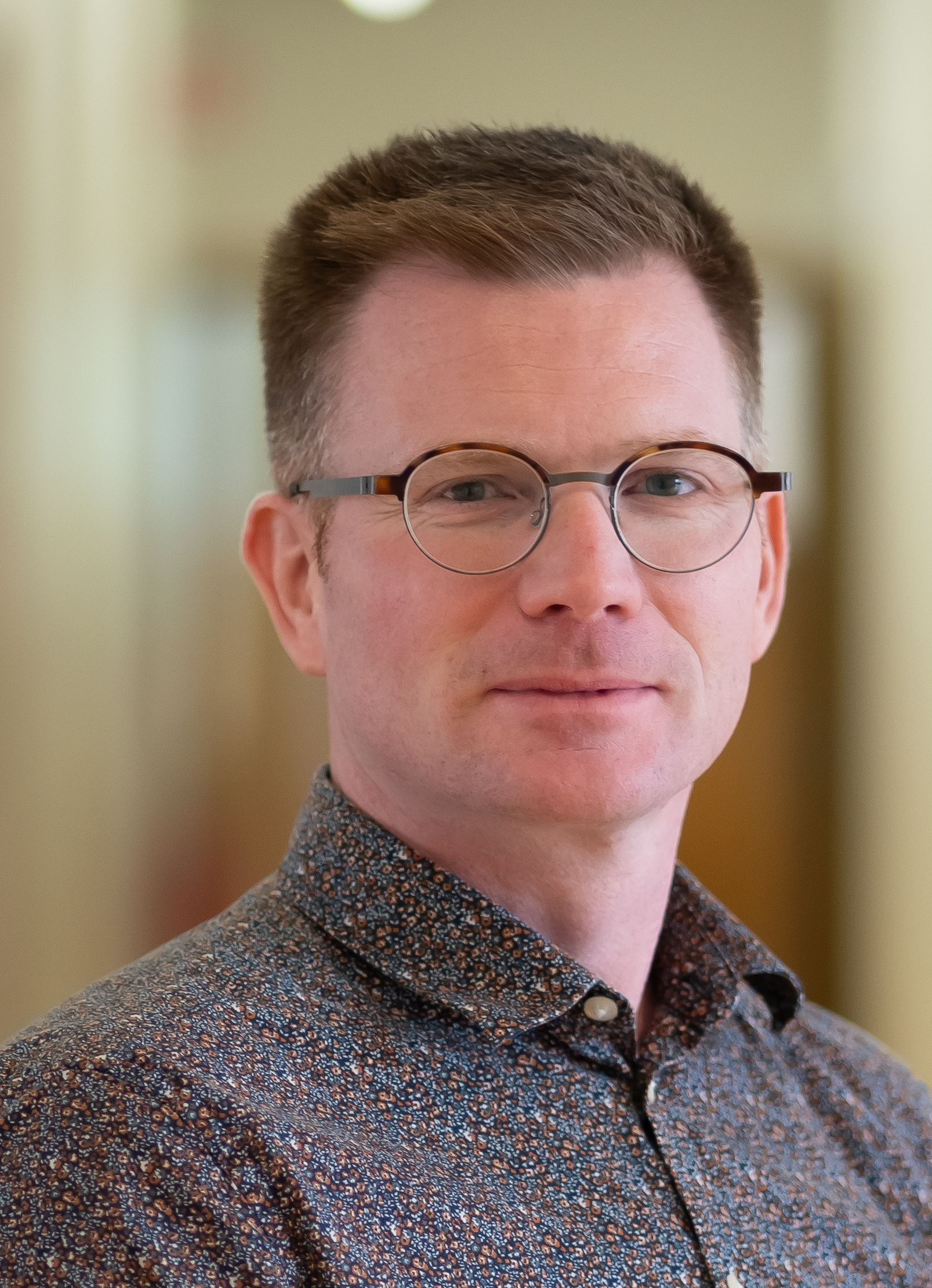}}]{Pieter Simoens} currently holds a position as Associate Professor at Ghent University, and is also affiliated with imec.

His main research interests include the domain of intelligent distributed systems, with a specific focus on resource-efficiency, unsupervised learning and collective intelligence.
\end{IEEEbiography}

\end{document}